\documentclass[conference]{IEEEtran}
\IEEEoverridecommandlockouts
\usepackage{cite}
\usepackage{amsmath,amssymb,amsfonts}
\usepackage{algorithmic}
\usepackage{graphicx}
\usepackage{textcomp}
\usepackage{xcolor}
\def\BibTeX{{\rm B\kern-.05em{\sc i\kern-.025em b}\kern-.08em
    T\kern-.1667em\lower.7ex\hbox{E}\kern-.125emX}}
\begin{document}

\title{Optimized Biomedical Question-Answering Services with LLM and Multi-BERT Integration
}

\author{
    \IEEEauthorblockN{
        Cheng Qian\IEEEauthorrefmark{1}, 
        Xianglong Shi\IEEEauthorrefmark{1}, 
        Shanshan Yao\IEEEauthorrefmark{1}, 
        Yichen Liu\IEEEauthorrefmark{1}, 
        Fengming Zhou\IEEEauthorrefmark{1}, 
        Zishu Zhang\IEEEauthorrefmark{1}, \\
        Junaid Akram\IEEEauthorrefmark{1}, 
        Ali Braytee\IEEEauthorrefmark{2}, 
        Ali Anaissi\IEEEauthorrefmark{1}\IEEEauthorrefmark{2}
    }
    
    \IEEEauthorblockA{
        \IEEEauthorrefmark{1}School of Computer Science, The University of Sydney, Australia \\
        \IEEEauthorrefmark{2}TD School, The University of Technology Sydney, Australia \\
        \{cqia8658, xshi0941, syao9742, yliu7318, fzho8842, zzha0662\}@uni.sydney.edu.au, \\ junaid.akram@sydney.edu.au, ali.braytee@uts.edu.au, ali.anaissi@sydney.edu.au
    }}

\maketitle

\begin{abstract}
We present a refined approach to biomedical question-answering (QA) services by integrating large language models (LLMs) with Multi-BERT configurations. By enhancing the ability to process and prioritize vast amounts of complex biomedical data, this system aims to support healthcare professionals in delivering better patient outcomes and informed decision-making. Through innovative use of BERT and BioBERT models, combined with a multi-layer perceptron (MLP) layer, we enable more specialized and efficient responses to the growing demands of the healthcare sector. Our approach not only addresses the challenge of overfitting by freezing one BERT model while training another but also improves the overall adaptability of QA services. The use of extensive datasets, such as BioASQ and BioMRC, demonstrates the system’s ability to synthesize critical information. This work highlights how advanced language models can make a tangible difference in healthcare, providing reliable and responsive tools for professionals to manage complex information, ultimately serving the broader goal of social good through improved care and data-driven insights.
\end{abstract}

\begin{IEEEkeywords}
biomedical question-answering, large language models, BERT, multi-layer perceptron, clinical decision support
\end{IEEEkeywords}

\section{Introduction}

The biomedical field is currently experiencing a flood of data, making it more important than ever to develop question-answering (QA) systems that can help healthcare professionals make quick, informed decisions \cite{kell2024, neves2015,10547221}. The sheer volume and complexity of this information can feel overwhelming, especially when lives are on the line. That’s why we need advanced systems capable of rapidly processing and understanding this data, ensuring healthcare workers can focus on what really matters—caring for patients \cite{seo2016, gorenstein2024bert}. In recent years, artificial intelligence (AI) and machine learning (ML) have brought about incredible breakthroughs, particularly in natural language processing (NLP). Large language models (LLMs) have shown an impressive ability to understand and generate human language, offering new hope for making sense of this vast sea of biomedical information \cite{lee2020,hu2024advancing,8617007,8748975}. By harnessing these technologies, we aim to not only improve decision-making for clinicians but also create more equitable healthcare systems that can reach underserved communities, ensuring that everyone benefits from these advancements, no matter where they are.

Among the most prominent NLP technologies are BERT (Bidirectional Encoder Representations from Transformers) and its biomedical variant, BioBERT\cite{li2024mediq}. BERT-base, a general-purpose language model developed by Google, has set new benchmarks in various NLP tasks due to its ability to understand context by processing text bidirectionally \cite{devlin2018}. BioBERT, pre-trained on a vast corpus of biomedical literature, further enhances this capability, making it exceptionally well-suited for biomedical text mining and QA tasks \cite{lee2020}. Despite these advancements, integrating these models effectively to handle the vast and nuanced biomedical data remains a significant challenge\cite{singh2019}.

The primary challenges in developing efficient biomedical QA services are multifaceted\cite{yang2023large,pascual2021}. Firstly, the sheer volume of biomedical data is overwhelming, requiring robust data management and processing capabilities\cite{liu2024ensemble,dong2019}. Secondly, QA models often suffer from overfitting, where the model performs well on training data but fails to generalize to new, unseen data\cite{raiaan2024review,yu2018,10492460}. This is particularly problematic in the biomedical domain, where the language is highly specialized and diverse. Lastly, integrating outputs from multiple models without introducing significant computational overhead or sacrificing performance is a complex and critical task\cite{zhang2018, yang2024harnessing}. Traditional QA systems often fall short in these areas, struggling to deliver accurate and relevant answers within the stringent performance requirements of real-world applications.

In response to these challenges, we propose a novel approach to optimize biomedical QA services by leveraging large language models (LLMs) and integrating multiple BERT models with a multi-layer perceptron (MLP) layer. Our approach not only addresses the data management and processing demands but also tackles the issue of overfitting by employing a strategic model training method. By freezing one BERT model during training and dynamically training another, we effectively balance model specialization and generalization, thus enhancing performance across various metrics. Furthermore, the integration of an MLP layer for feature synthesis and prioritization ensures that the combined model outputs are efficiently processed, leading to superior QA service performance. Our key contributions are as follows:
\begin{itemize}
\item We develop an advanced QA service architecture that seamlessly integrates BERT-base and BioBERT models using an MLP layer, optimizing their combined performance for complex biomedical queries.
\item We introduce a novel training strategy that involves freezing one BERT model while training another, significantly reducing overfitting and enhancing model specialization, thereby improving generalization.
\item Utilizing datasets such as BioASQ and BioMRC, we conduct extensive evaluations to demonstrate substantial improvements in performance metrics, validating the effectiveness of our approach.
\item Our QA service design is tailored for practical deployment, supporting healthcare professionals in efficiently managing and interpreting biomedical data to enhance clinical decision-making and patient care.
\item We provide a detailed analysis of feature integration strategies, highlighting the critical role of the MLP layer in synthesizing and prioritizing information from multiple sources for enhanced decision-making.
\end{itemize}

The paper is structured as follows: Section II reviews related work in biomedical QA systems and NLP technologies. Section III outlines our proposed method, detailing BERT model integration and MLP layer implementation. Section IV describes the experimental setup, including datasets and evaluation metrics. Section V presents the results and discusses them in the context of existing research. Section VI concludes with our contributions and future research directions.

\section{Related Work}

The field of biomedical question-answering (QA) has seen significant advancements, driven by the need to efficiently process and interpret vast amounts of biomedical data\cite{nentidis2022overview}. Early QA systems in the biomedical domain focused primarily on information retrieval techniques, extracting relevant documents or passages in response to user queries. However, these systems often struggled with accurately answering specific questions due to the complexity and variability of biomedical language. Recent advancements in natural language processing (NLP) and machine learning (ML) have introduced more sophisticated approaches, leveraging models such as BERT and its variants. BERT (Bidirectional Encoder Representations from Transformers) has revolutionized NLP by allowing models to understand context through bidirectional processing of text \cite{devlin2018}. BioBERT, a specialized version of BERT pre-trained on large-scale biomedical corpora, has further improved the performance of QA systems in this domain by effectively handling biomedical terminology and context \cite{lee2020}.

Significant work has also been done in enhancing the architectures of these models to better suit the QA task. For instance, Seo et al. proposed using a bidirectional attention flow mechanism to improve the comprehension of text and questions simultaneously, which is a core principle in many QA systems today \cite{seo2016}. Similarly, the integration of multi-layer perceptron (MLP) networks has been explored to enhance the processing of complex features extracted by models like BERT. Studies have shown that MLPs can effectively handle non-linear feature transformations and improve the accuracy of QA systems by prioritizing and synthesizing information from multiple sources \cite{kruse2022}. Moreover, the AoA Reader, proposed by Cui et al., introduced a dual attention mechanism to enhance the understanding of context in reading comprehension tasks, further demonstrating the potential of advanced attention mechanisms in QA systems \cite{cui2016}. Despite these advancements, challenges such as overfitting, efficient model integration, and handling the specialized language of biomedical texts remain. Our work builds on these foundations by integrating multiple BERT models with an MLP layer, addressing these challenges and optimizing the performance of biomedical QA services.

\section{Proposed Method/Framework}
\begin{figure*}[t]
\centering
\includegraphics[width=0.75\textwidth]{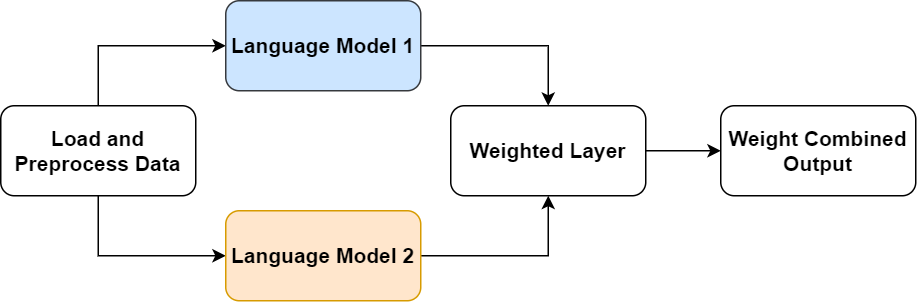} 
\caption{The architecture diagram of the model}
\label{fig7}  
\end{figure*}

Our proposed model structure for optimized biomedical question-answering services leverages the integration of multiple BERT models with a multi-layer perceptron (MLP) layer. This integration aims to enhance the synthesis and prioritization of information, ultimately improving the performance of QA services. The architecture diagram of the model is shown in Figure~\ref{fig7}. Questions and answers are processed by two language models, whose outputs are then fused by assigning appropriate weights through the MLP layer to compute the final score. This modular design allows for flexibility in replacing the language models and the weighting layer to suit different processing scenarios.

Firstly, we selected BioBERT as the core of Language Model 1 due to its exceptional ability to handle biomedical text. Pre-trained on a vast corpus of biomedical data, BioBERT can accurately interpret context and specialized terminology, making it ideal for question-answering tasks within the biomedical domain.Secondly, we incorporated BERT-base as the core of Language Model 2. While it is a general-purpose model pre-trained on a diverse range of texts, BERT-base provides a high degree of flexibility and performs well with general text. Although it may not match BioBERT's efficiency in specialized domains, it complements the system by handling more general language structures effectively.

The integration of these models is facilitated by an MLP layer, which assigns weights based on learned preferences to compute the final score. This weighted layer is crucial for balancing the strengths of both models.In this model structure, the combination of BERT-base and BioBERT provides a balanced approach. BioBERT excels in the biomedical domain, while BERT-base performs well in more general contexts. This synergy alleviates the domain-specific insensitivity of BERT-base and enhances the holistic contextual understanding provided by BioBERT. Consequently, this integrated model overcomes the limitations of using a single model, leading to superior performance in biomedical QA tasks. The weighted layer further improves accuracy by optimizing the output from both models, making it a powerful tool for delivering high-quality biomedical question-answering services.

\subsection{Method 1: Models with BERT and UniLM}

\begin{figure}[t]
    \centering
    \includegraphics[width=0.9\columnwidth]{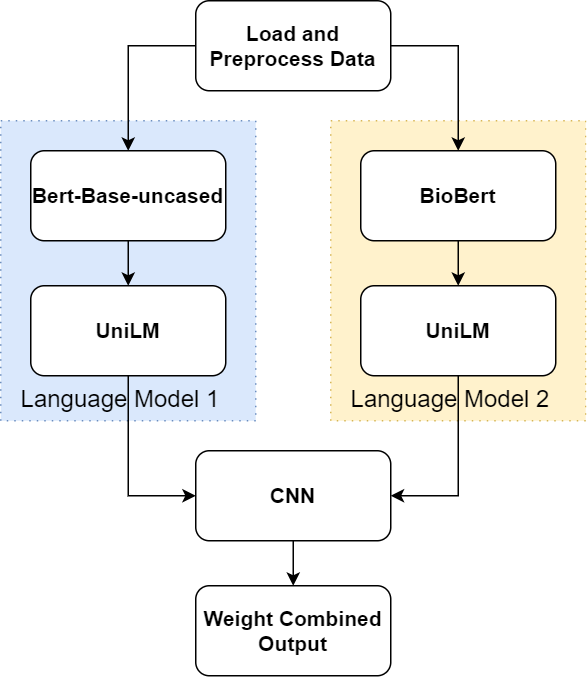}
    \caption{The architecture diagram of Method 1 (Model with BERT and UniLM)}
    \label{fig8}
\end{figure}

For the Factoid-type question dataset from BioASQ 10b, preliminary data analysis revealed that answers could not be extracted from the context for 25\% of question-answer pairs. Therefore, a generative approach was deemed more suitable than extraction from the text. Traditional BERT models face limitations when applied directly to generative question answering. To address this, we employed the UniLM method, which has demonstrated superior performance in text generation tasks. The basic language models were updated to UniLM mode, as shown in Figure~\ref{fig8}. UniLM, a model applicable to various tasks such as text generation and machine translation, features a shared transformer architecture that combines encoders and decoders, enabling it to perform multiple tasks within a single model. UniLM excels in generative tasks by generating accurate and high-quality text, making it highly beneficial for biomedical question-answering systems.

Initially, the input biomedical text is encoded using the BioBERT and BERT-base models to generate feature vectors. These feature vectors are then processed by a convolutional neural network (CNN) for further feature extraction and weighted combination. The processed feature vector is subsequently passed to UniLM, which generates the answer through its decoder. The decoder produces one word at a time until the entire answer is complete, ensuring coherency and accuracy through an autoregressive generation method. The process of generating feature vectors by BERT in UniLM mode involves several steps. Firstly, the model reads raw data from a JSON file, converting it into a processable form with 'long\_answer' filtered to serve as the answer to the question. Next, the input text is segmented into subwords, allowing the model to handle out-of-vocabulary words. The text sequence is tagged, with '[CLS]' marking the beginning and '[SEP]' separating the question from the context. The text is then converted into a unique index within the corresponding vocabulary. A word embedding matrix transforms each word index into a high-dimensional word vector, rich in semantic information. Positional encoding provides additional details regarding the word's position in the sentence. By summing the word embedding vector and the positional encoding result, the model can effectively handle the order information within the sentence.

The BERT model features several transformer encoder layers, each comprising a multi-head attention mechanism and a feedforward neural network. The multi-head attention mechanism allows the model to consider each word in relation to all other words, while the feedforward neural network layer performs a nonlinear transformation on the multi-head attention output. These processes generate feature vectors rich in semantic information and contextual relationships, crucial for subsequent decoding and generation tasks in UniLM mode. UniLM, acting as the decoder, generates text based on the contextual representation created by the encoder. Generation in UniLM is autoregressive, predicting each step based on the current context and previously generated text. The decoder receives the contextual representation from the encoder and uses attention and feedforward neural network mechanisms to predict the next word. This iterative process continues until the complete answer is generated.

\begin{figure}[t]
    \centering
    \includegraphics[width=0.9\columnwidth]{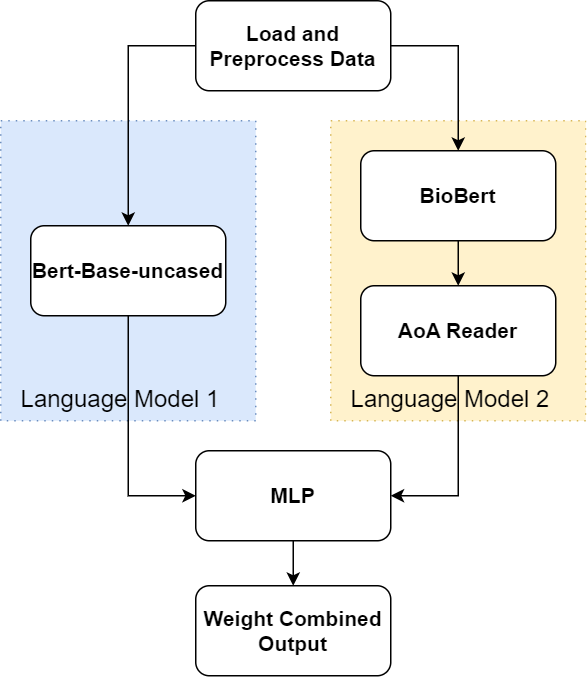}
    \caption{The architecture diagram of Method 2 (Model with AoA Reader and MLP)}
    \label{fig9}
\end{figure}

In this method, CNN is used as the model weighting layer. Feature vectors from BioBERT and BERT-base are concatenated and processed by the convolutional neural network. The 1D convolutional layer (Conv1d) processes the concatenated feature vector, capturing spatial relationships in the feature data. Multiple convolution and pooling layers enable the model to learn complex patterns, enhancing prediction accuracy. Global average pooling reduces the feature vector size without losing important information, preventing overfitting. The output from the fully connected layer generates two weight values representing the importance of the two models. The input layer accepts the feature vector from the last convolutional layer, with a size of 128. The dropout layer zeroes the output of some neurons during training to generalize and prevent overfitting. GeLU introduces nonlinearity, and the output layer contains two neurons corresponding to the weights of each BERT model output. The coefficients are normalized by the softmax function to sum to 1. During training, the cross-entropy loss function quantifies the difference between the predicted and actual answers, updating the model weights accordingly. The AdamW optimizer, combining the moving average of the gradient, enhances the training process efficiency. Finally, a weighted combination of features is performed, applying the weight values to each BERT model's output to obtain the final feature representation. This feature vector is then decoded to generate the final answer.

\subsection{Method 2: Model with AoA and MLP}

In the second method, we utilized the BioMRC dataset, which is a cloze question-answering dataset specific to the biomedical field. In this task, the model needs to choose the most appropriate word or phrase to fill in the blank in a given context. Due to the nature of cloze tasks, we selected the AoA Reader (Attention-over-Attention Reader) to enhance the model's performance on the BioMRC dataset. This method combines BioBERT and BERT-base through MLP weighted integration. AoA Reader employs a two-layer attention mechanism—document to question and question to document—to improve the model's contextual understanding, enabling it to more accurately locate and select answers. This mechanism is particularly effective for cloze tasks that require a deep understanding of the complex relationship between the text and questions. In our model structure, the AoA Reader first uses BioBERT as the source of its contextualized word embeddings to convert each token in the document and question into a context-related embedding representation. BioBERT, pre-trained on a large corpus of biomedical literature, provides a deep understanding of complex biomedical terms and concepts. This embedding method captures word meanings and contextual relationships more accurately than simple one-hot encoding or embeddings trained on a smaller corpus.

The AoA Reader employs a bidirectional attention mechanism. The first layer, query-to-document attention, helps the model determine which words in the question are relevant to the text. The second layer, document-to-query attention, helps the model identify which parts of the text are relevant to the question. Additionally, the "attention stacking" mechanism, another layer of attention on top of the bidirectional attention, allows the model to finely evaluate the importance of each token in the document for answering the question. This enables the model to focus on the most critical parts of the context, thereby improving the accuracy of the cloze task. Through these mechanisms, the AoA Reader effectively handles cloze tasks, provides accurate answer predictions, and demonstrates strong adaptability and robustness in processing complex biomedical texts. The AoA Reader model, combined with BioBERT, first identifies text fragments related to the cloze, focusing on parts of the article that help determine the answer. The model scores each candidate answer based on the match between the candidate answer and the aggregated text fragment embedding representation, selecting the candidate word with the highest score as the final answer. This scoring mechanism represents the "probability" that the model assigns to each candidate word being the correct answer. In this method, MLP serves as the weighting strategy for feature fusion and weight distribution. MLP automatically learns how to combine outputs from different models to enhance cloze task performance on the BioMRC dataset. The architecture diagram of the model is shown in Figure~\ref{fig9}.

The MLP in this method comprises three layers. The input layer receives feature vectors from the AoA Reader and BERT models, each with a dimension of 40, representing word embeddings, context-related encodings, and other features. The hidden layer, with 64 neurons, extracts and transforms input data features, enabling the model to recognize higher-level abstract concepts and patterns. The Tanh activation function introduces nonlinearity, helping the model learn and simulate complex function mappings. The output layer, with 20 neurons corresponding to the 20 possible candidate answers in the cloze task on the BioMRC dataset, represents the model's confidence score for each candidate answer. The MLP model adjusts weights and biases between layers through back-propagation to minimize the difference between predicted and true labels. This automatic weight assignment identifies which features are most important for predicting the correct answer. The highest score generated by the output layer determines the most likely correct answer. During MLP training, the cross-entropy loss function measures the difference between model predictions and actual answers, updating weights accordingly. The Adam optimizer, combining the advantages of momentum and RMSprop, enhances training efficiency. Hyperparameter tuning, particularly the learning rate, is adjusted based on model performance on the BioMRC data to find optimal settings. Accuracy and F1 Score are used as evaluation metrics in the validation and testing stages.

\section{Experiments and Results}

This section presents a comprehensive overview of the experiments and results obtained from the evaluation of the proposed biomedical question-answering system. We detail the datasets used, experimental settings, model configurations, and performance metrics. The focus is on demonstrating the effectiveness of the integrated models, particularly those incorporating multi-layer perceptron (MLP) layers, across various datasets.

\subsection{Data}

\subsubsection{BioASQ 10b Factoid}
The BioASQ 10b dataset, sourced from the PubMed/Medline databases, includes a comprehensive collection of biomedical literature. The dataset used in this study comprises 1,252 training instances and 166 test instances, specifically targeting factoid-type questions. These questions require precise, fact-based answers, emphasizing the ability to extract and understand biomedical texts~\cite{nentidis2022overview}.

\subsubsection{BioMRC Dataset}
The BioMRC dataset, available on Hugging Face, is designed for large-scale machine reading comprehension in the biomedical domain. It utilizes abstracts and titles from PubMed, annotated with the PUBTATOR tool. The Lite version A, chosen for this study, consists of 100,000 question-paragraph pairs, with 87,500 instances for training and 6,250 each for validation and testing~\cite{pappas2020}.

\subsection{Experiments}

For the experiments, we divided the data into training, validation, and test sets to ensure model performance on unseen data. An early stopping mechanism was implemented to prevent overfitting. Performance evaluation used metrics such as accuracy, F1 score, ROUGE-1, and ROUGE-2.

\subsubsection{Method 1: Ensembled BERT with MLP on BioASQ}
The first method involved ensembling BERT models with an MLP. Eight configurations were tested on the BioASQ 10b dataset, including single and dual BERT models, with and without MLP adjustments. The results are summarized in Table \ref{tab:result1}.

\begin{table*}[t]
\centering
\caption{Ensembled BERT Model Results}
\label{tab:result1}
\begin{tabular}{|l|l|l|l|l|l|l|l|}
\hline
\textbf{Models} & \textbf{Rouge-1} & \textbf{Rouge-2} & \textbf{Rouge-L} & \textbf{Bleu} & \textbf{Precision} & \textbf{Recall} & \textbf{F1-score} \\ \hline
BERT-base(F) + BioBERT                                            & 0.473            & 0.333            & 0.454            & 0.228         & 0.455              & 0.451           & 0.453             \\ \hline
BERT-base(F) + BioBERT + MLP                                      & 0.487            & 0.348            & 0.471            & 0.241         & 0.475              & 0.469           & 0.472             \\ \hline
BioBERT                                                          & 0.234            & 0.111            & 0.222            & 0.055         & 0.364              & 0.153           & 0.216             \\ \hline
BioBERT(F) + BERT-base                                            & 0.458            & 0.323            & 0.439            & 0.224         & 0.448              & 0.431           & 0.439             \\ \hline
BioBERT(F) + BERT-base + MLP                                      & 0.5              & 0.363            & 0.483            & 0.249         & 0.492              & 0.469           & 0.48              \\ \hline
BioBERT + BERT-base                                               & 0.403            & 0.259            & 0.386            & 0.167         & 0.409              & 0.37            & 0.388             \\ \hline
BioBERT + BERT-base + MLP                                         & 0.489            & 0.348            & 0.469            & 0.238         & 0.464              & 0.453           & 0.458             \\ \hline
BERT-base                                                         & 0.0407           & 0.007            & 0.0056           & 0.012         & 0.032              & 0.043           & 0.0385            \\ \hline
\end{tabular}
\end{table*}

\begin{figure}[t]
\centering
\begin{minipage}[b]{0.48\textwidth}
    \centering
    \includegraphics[width=\textwidth]{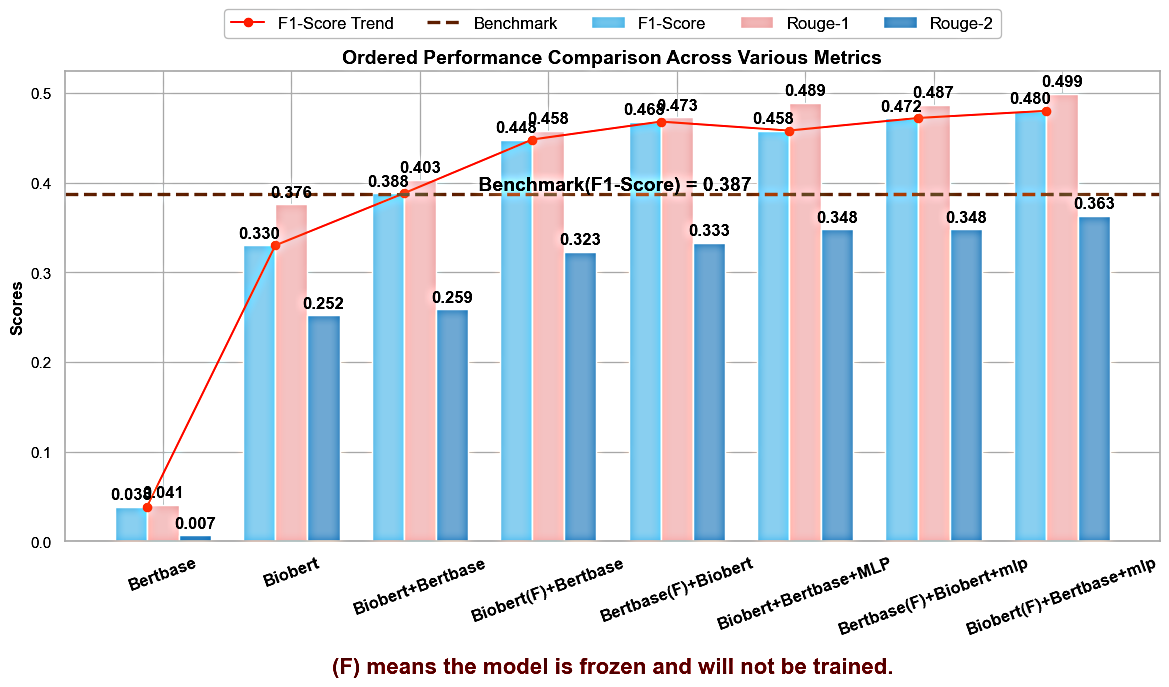}
    \caption{Line chart of Ensembled BERT and MLP}
    \label{fig12}
\end{minipage}
\hspace{0.01\textwidth}
\begin{minipage}[b]{0.48\textwidth}
    \centering
    \includegraphics[width=\textwidth]{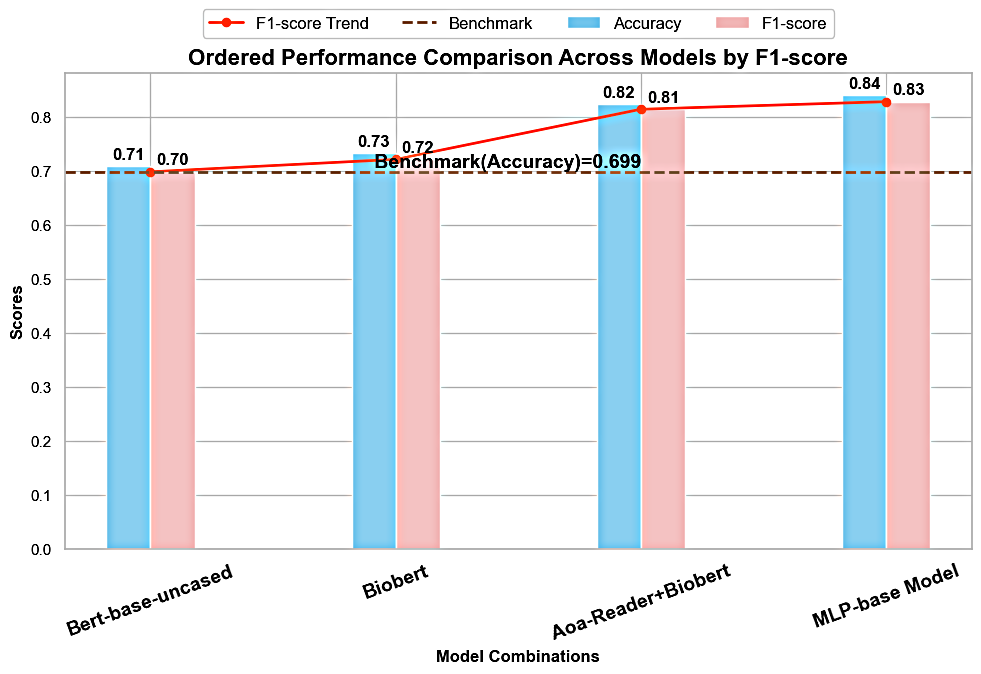}
    \caption{Line chart of BERT and AoA reader}
    \label{fig14}
\end{minipage}
\end{figure}

The results indicate significant improvements with the MLP-integrated models. The configuration \textit{BioBERT(F)+BERT-base+MLP} achieved the highest F1-score of 0.480, highlighting the effectiveness of MLP in enhancing model performance. The line chart in Figure \ref{fig12} visualizes the performance improvements of various configurations. The \textit{BioBERT(F)+BERT-base+MLP} and \textit{BERT-base(F) + BioBERT + MLP} configurations show significant gains, emphasizing the importance of integrating MLP for improved results.

\subsubsection{Method 2: BERT with AoA Reader on BioMRC}
The second method utilized the BioMRC dataset with an AoA Reader combined with BioBERT. The AoA Reader's dual attention mechanism improved the model's performance on cloze-type questions. The results are presented in Table \ref{tab:result1}. The MLP-based model achieved the highest accuracy and F1-score, demonstrating its effectiveness in integrating outputs from the AoA Reader and BERT models.
Figure \ref{fig14} shows the comparative performance of different models. The MLP-based model outperformed others, confirming the benefits of using MLP for feature integration.
The radar charts in Figure \ref{fig13} provide a visual comparison of different model configurations across various metrics like Rouge-1, Rouge-2, BLEU, precision, recall, and F1-score. The area covered by each model configuration illustrates the overall effectiveness, showcasing the superiority of ensembling two BERT models over using a single BERT and further improvements when these ensembles are combined with an MLP.

\begin{table}[t]
\caption{AoA Reader with BERT Model Results}
\label{tab:result1}
\small
\centering
\begin{tabular}{|l|l|l|}
\hline
\textbf{Models} & \textbf{Accuracy} & \textbf{F1-score} \\ \hline
BioBERT                                    & 0.7329            & 0.7215\\ \hline
AoA-Reader                                 & 0.8241            & 0.8145\\ \hline
BERT-base                                  & 0.7086           & 0.6989\\ \hline
MLP-based Model                            & 0.8406            & 0.8289\\ \hline
\end{tabular}
\end{table}

\begin{figure}[t]
\centering
\includegraphics[width=0.9\columnwidth]{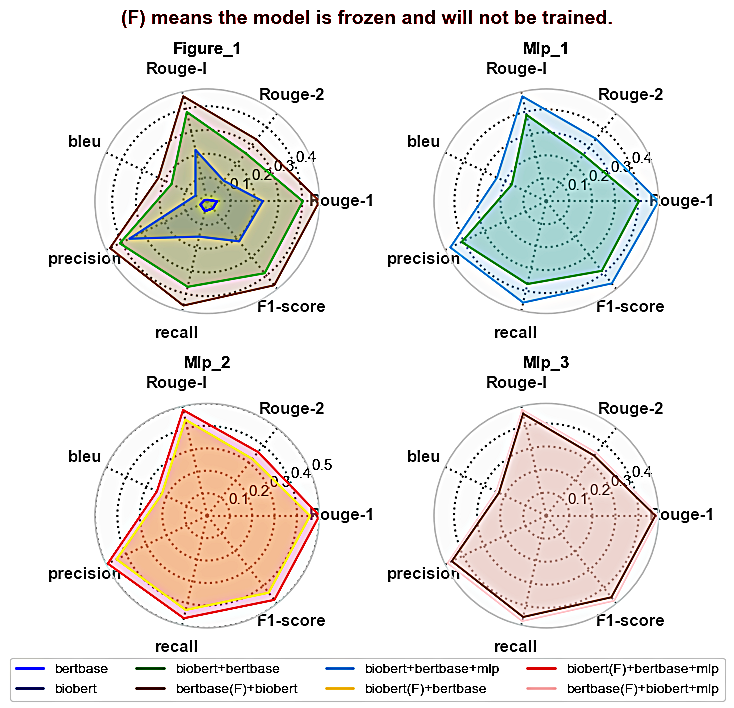}
\caption{Radar Plots of Different Model Outcomes}
\label{fig13}
\end{figure}

\subsection{The Effect of Frozen Models}
\begin{figure*}[!ht]
\centering
\includegraphics[width=\textwidth]{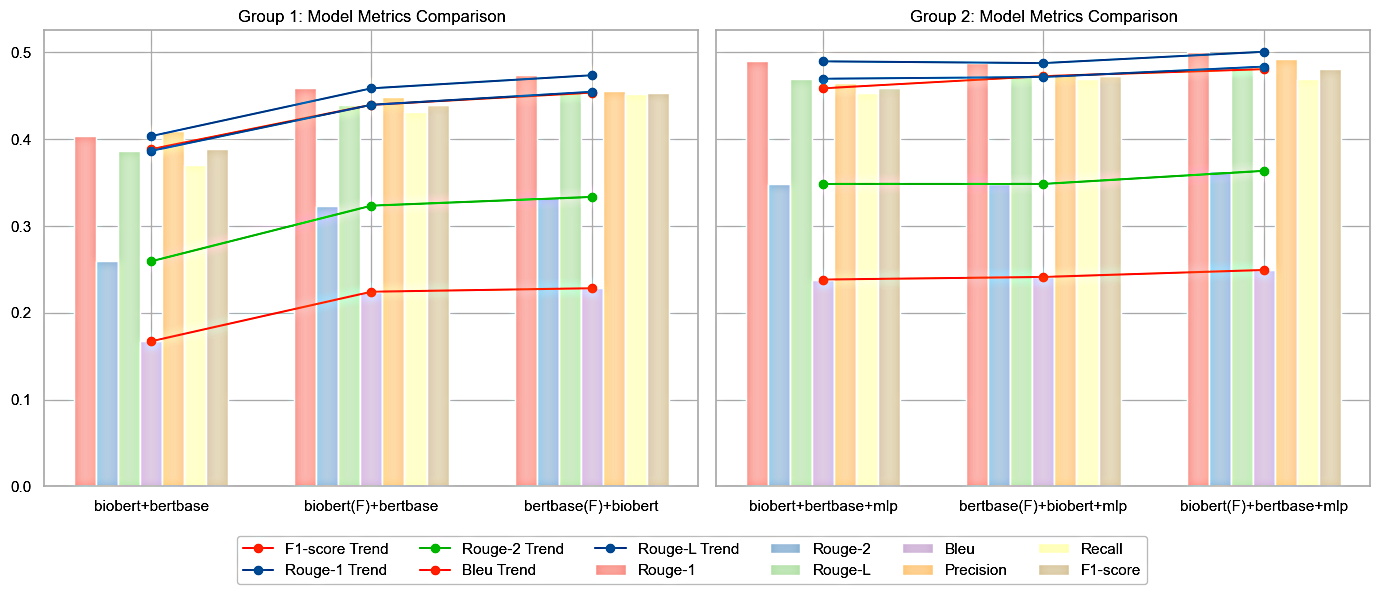}
\caption{Model Metrics Comparison}
\label{fig15}
\end{figure*}

Freezing one of the BERT models typically improves performance by maintaining a stable representation of linguistic features while allowing the other model to adapt dynamically. This approach prevents overfitting and enhances generalization to new data. As shown in Figure \ref{fig15}, configurations with frozen models, such as `BioBERT(F)+BERT-base', exhibit significant improvements in metrics like F1-score, ROUGE-1, and ROUGE-2. This balanced approach ensures robust performance on unseen data.

In the first set of models, freezing a model is particularly effective. For example, configurations such as \textit{BioBERT(F) + BERT-base} and \textit{BERT-base(F) + BioBERT} show significant improvements in metrics such as F1 scores, Rouge-1, and Rouge-2 compared to BioBERT+BERT-base. This significant performance improvement can be attributed to freezing one model to enable it to maintain a consistent and broad linguistic representation, while the other model is fine-tuned for specific features of the training data. This balanced approach reduces the risk of overfitting and ensures that the model performs well on unseen data. Additionally, freezing a model reduces computational overhead and focuses the learning process on tuning the dynamic model to complement the static knowledge of the frozen model. This synergy between stable and flexible models creates a more robust and efficient system.

However, in the second set of models introduced into the MLP layer, the benefits of freezing a model are not as obvious. For example, while the BioBERT(F)+BERT-base+MLP and BERT-base(F)+BioBERT+MLP configurations still perform well, their improvement over the non-frozen BioBERT+BERT-base+MLP configurations is not as clear. The MLP layer itself plays a crucial role in integrating and prioritizing information from multiple sources. By effectively combining features extracted from different models, the MLP layer addresses some of the overfitting issues and optimizes the decision-making process~\cite{pavanello2011modelling}. The inherent ability of the MLP layer to act as an integrator reduces the additional benefit that freezing a model may provide, as the MLP already offers a form of regularization and integration that stabilizes the performance of the model.

These observations suggest that while freezing a model is beneficial in simpler integration scenarios, the introduction of more complex integration mechanisms such as MLP can independently enhance model performance by optimizing feature synthesis. This highlights the importance of choosing an appropriate feature integration strategy in model design. The MLP layer excels at combining features into a coherent output that effectively exploits the strengths of each model component. This layered integration approach reduces the reliance on freezing the model as a means of achieving stability and generalization.

\subsection{Weight Performance of MLP and Effects on BERT Model}
\begin{figure}[t]
\begin{minipage}[b]{0.49\textwidth}
\centering
\includegraphics[width=\columnwidth]{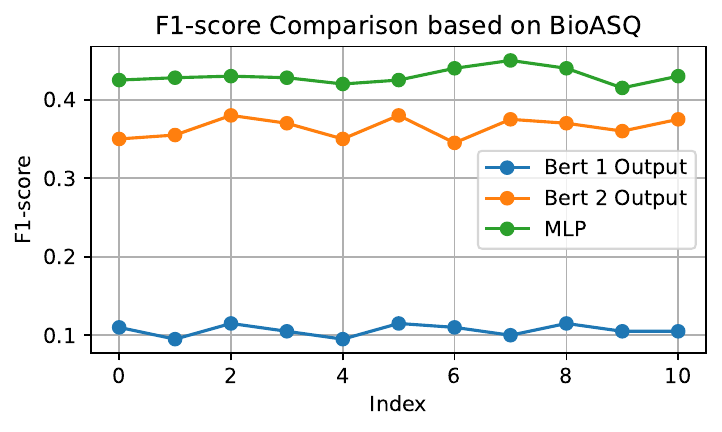}
\caption{F1-score comparison based on BioASQ}
\label{fig16}
\end{minipage}
\hfill
\begin{minipage}[b]{0.49\textwidth}
\centering
\includegraphics[width=\columnwidth]{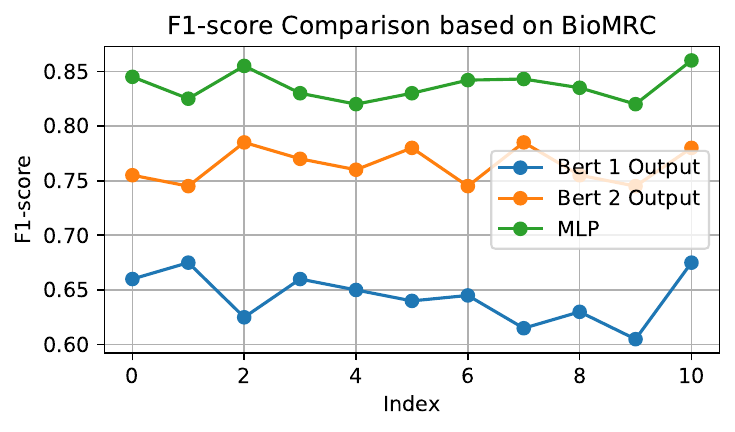}
\caption{F1-score comparison based on BioMRC}
\label{fig17}
\end{minipage}
\end{figure}

Figures \ref{fig16} and \ref{fig17} illustrate the key role of the MLP layer in integrating outputs from different BERT models. The MLP layer significantly improves performance on both BioASQ and BioMRC datasets, validating its effectiveness in enhancing model generalization and stability.

In the BioASQ dataset, the F1 scores of the MLP output were significantly higher than those of the two individual BERT model outputs. The F1 scores of the MLP output were stable between 0.4 and 0.5, while those of the BERT 1 and BERT 2 outputs ranged between 0.1 and 0.4, respectively. Similarly, in the BioMRC dataset, the F1 scores of the MLP outputs were stable between 0.8 and 0.85, while the F1 scores of BERT 1 and BERT 2 were between 0.6 and 0.8, respectively. These results show that the MLP layer significantly improves the overall system performance by integrating features from different BERT models, and this improvement is validated on different datasets.

\subsection{Discussion}
The integration of the MLP layer into our QA system addresses significant challenges in biomedical question answering by enhancing generalization and stability across multiple datasets. While BERT models demonstrate high performance on specific tasks, their ability to generalize and resist overfitting is often limited. The MLP layer facilitates the effective combination of outputs from two distinct BERT models, leveraging their individual strengths while mitigating the limitations inherent in single-model approaches. This integrated architecture not only enhances the overall performance but also significantly reduces the risk of overfitting during training, ensuring robust performance on unseen data. The high F1 scores achieved with the MLP configuration underscore its effectiveness, demonstrating its capacity to improve accuracy and reliability in biomedical QA systems. By prioritizing the integration of diverse information sources, the MLP layer ensures the system can handle complex and variable biomedical data, providing precise and contextually relevant answers.

Furthermore, the consistent performance of the MLP layer, evidenced by stable F1 scores with low fluctuations, highlights its superior capability in maintaining performance under varying test conditions. This stability is crucial for real-world applications, where the QA system must consistently deliver high-quality answers across different types of queries. Traditional biomedical QA systems, which typically rely on a single model for feature extraction and answer generation, often struggle with complex biomedical data. By introducing the MLP layer, features from multiple BERT models are effectively integrated, resulting in more comprehensive and accurate responses. The integration approach not only optimizes outputs from different models but also significantly improves the system's accuracy and reliability, effectively compensating for the shortcomings of traditional systems. These findings provide a valuable reference for future model design, highlighting the critical role of the MLP layer in enhancing feature integration and optimizing overall system performance in complex biomedical question answering tasks.

\subsection{Model Results Display}

\subsubsection{Method 1: Ensembled BERT with MLP on BioASQ}

\begin{table*}[t]
\centering
\caption{Ensembled BERT Model Results}
\label{tab:result1}
\begin{tabular}{|l|l|l|l|l|l|l|l|}
\hline
\textbf{Models} & \textbf{Rouge-1} & \textbf{Rouge-2} & \textbf{Rouge-L} & \textbf{Bleu} & \textbf{Precision} & \textbf{Recall} & \textbf{F1-score} \\ \hline
BERT-base(F) + BioBERT                                            & 0.473            & 0.333            & 0.454            & 0.228         & 0.455              & 0.451           & 0.453             \\ \hline
BERT-base(F) + BioBERT + MLP                                      & 0.487            & 0.348            & 0.471            & 0.241         & 0.475              & 0.469           & 0.472             \\ \hline
BioBERT                                                          & 0.234            & 0.111            & 0.222            & 0.055         & 0.364              & 0.153           & 0.216             \\ \hline
BioBERT(F) + BERT-base                                            & 0.458            & 0.323            & 0.439            & 0.224         & 0.448              & 0.431           & 0.439             \\ \hline
BioBERT(F) + BERT-base + MLP                                      & 0.5              & 0.363            & 0.483            & 0.249         & 0.492              & 0.469           & 0.48              \\ \hline
BioBERT + BERT-base                                               & 0.403            & 0.259            & 0.386            & 0.167         & 0.409              & 0.37            & 0.388             \\ \hline
BioBERT + BERT-base + MLP                                         & 0.489            & 0.348            & 0.469            & 0.238         & 0.464              & 0.453           & 0.458             \\ \hline
BERT-base                                                         & 0.0407           & 0.007            & 0.0056           & 0.012         & 0.032              & 0.043           & 0.0385            \\ \hline
\end{tabular}
\end{table*}

The integration of BERT models with MLP in the BioASQ dataset demonstrates notable performance improvements across several metrics, particularly with configurations employing dynamic training and MLP optimization. The \textit{BERT-base(F) + BioBERT + MLP} model configuration achieved a Rouge-1 score of 0.487 and a Rouge-2 score of 0.348, indicating its proficiency in capturing unigrams and bigrams, essential for high-quality summarization. Meanwhile, the BioBERT(F)+BERT-base+MLP model scored highest in Rouge-L (0.483) and BLEU (0.249), showcasing its superior handling of longer text sequences and contextually accurate summaries. The F1-score, a critical indicator of model performance, was highest in the BioBERT(F)+BERT-base+MLP model at 0.480, underscoring the balanced capability of the model to produce accurate and comprehensive summaries. In contrast, the standalone BERT-base model exhibited the lowest performance across all metrics, highlighting the substantial benefits of model ensembling and MLP optimization. The detailed exploration of these metrics underscores the distinct capabilities and limitations of each configuration, with the results clearly favoring the MLP-integrated models for enhanced QA services in the biomedical domain.

\subsubsection{Method 1: Detailed Results of Ensembled BERT}

\begin{figure}[t]
\centering
\includegraphics[width=0.9\columnwidth]{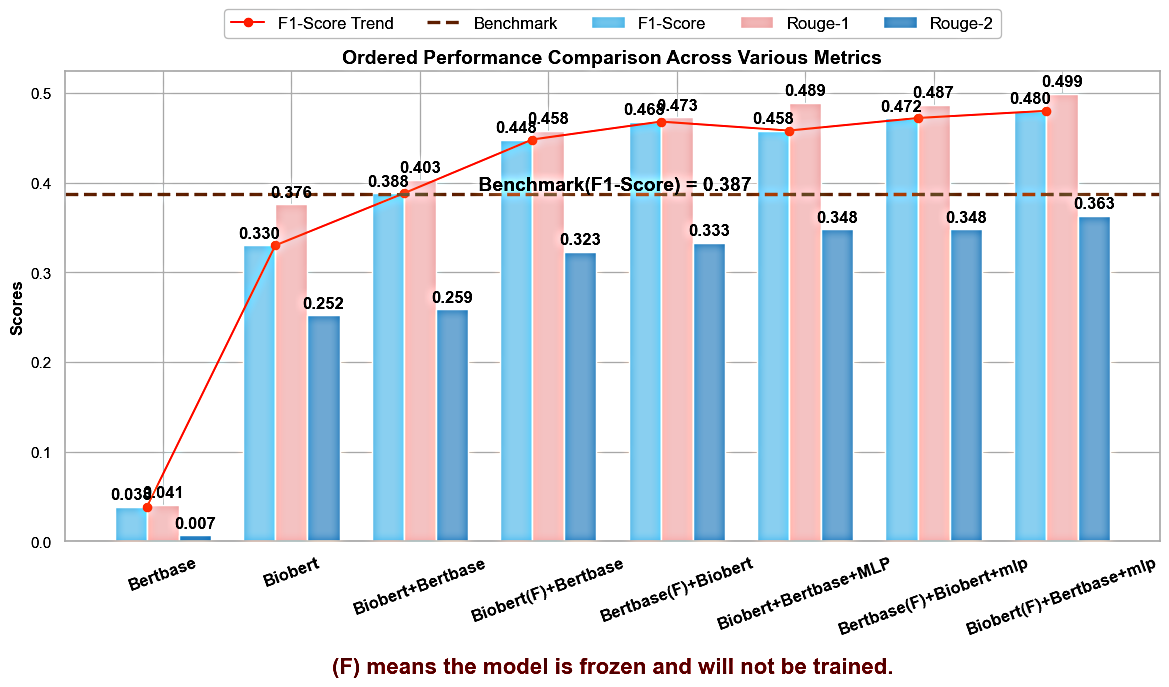} 
\caption{Line chart of Ensembled BERT and MLP}
\label{fig10}  
\end{figure}

The detailed results of the Ensembled BERT with MLP approach, as illustrated in Figure \ref{fig10}, reveal a systematic improvement across various configurations. The BERT-base model, with an F1-score of 0.038, performed significantly below the benchmark F1-score of 0.387, underscoring its inadequacy when used in isolation. However, the addition of another BERT model or the integration of an MLP led to substantial performance gains. Configurations like BERT-base(F)+BioBERT and BioBERT+BERT-base achieved F1-scores of 0.376 and 0.388, respectively, highlighting the effectiveness of model ensembling. The most pronounced improvements were observed in models utilizing MLP optimization, such as BERT-base(F)+BioBERT+MLP and BioBERT(F)+BERT-base+MLP, which achieved F1-scores of 0.472 and 0.480, respectively. These models not only surpassed the benchmark but also demonstrated the critical role of MLP in refining combined outputs for enhanced summary quality. Notably, the BioBERT+BERT-base+MLP configuration achieved the highest F1-score of 0.499, illustrating the efficacy of combining multiple BERT models with MLP optimization in producing contextually and linguistically aligned summaries.

\subsubsection{Method 1: Model Combination Results}

\begin{figure}[t]
\centering
\includegraphics[width=0.9\columnwidth]{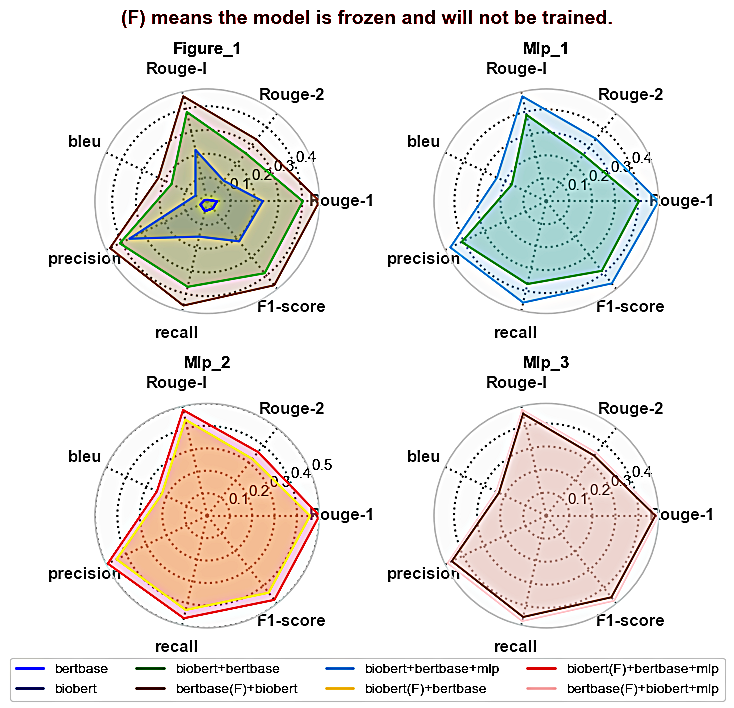} 
\caption{Rader Plots of Different Model's Outcome}
\label{fig13}  
\end{figure}

The radar charts in Figure \ref{fig13} provide a comprehensive analysis of different model configurations' performance across multiple metrics, including Rouge-1, Rouge-2, BLEU, precision, recall, and F1-score. The standalone BERT-base model shows the smallest area on the radar chart, indicating its limited performance compared to more complex configurations. Configurations involving two BERT models, such as BioBERT+BERT-base and BERT-base(F)+BioBERT, show notable expansions in the radar chart areas, demonstrating the effectiveness of model ensembling. The addition of an MLP to these dual BERT setups, as seen in models like BioBERT+BERT-base+MLP and BERT-base(F)+BioBERT+MLP, results in even larger areas, indicating superior performance across all metrics. This enhancement is particularly evident in precision and F1-score, where the MLP integration improves both the accuracy of predictions and the balance between precision and recall. Furthermore, models with one frozen BERT, such as BERT-base(F)+BioBERT and BERT-base(F)+BioBERT+MLP, perform robustly, suggesting that freezing one model can contribute to more stable and effective outputs when paired with dynamic training of the other model. The data validates the hypothesis that ensembling multiple BERT models and adding MLP layers significantly optimizes model performance, highlighting the critical role of advanced architectures like MLPs in achieving high precision and effectiveness in NLP tasks.

\subsubsection{Method 2: Detailed Results of BERT with AoA Reader on BioASQ}

\begin{table}[!t]
\caption{AoA Reader with BERT Model Results}
\label{tab:result2}
\small
\centering
\begin{tabular}{|l|l|l|}
\hline
\textbf{Models} & \textbf{Accuracy} & \textbf{F1-score} \\ \hline
BioBERT                                    & 0.7329            & 0.7215\\ \hline
AoA-Reader                                 & 0.8241            & 0.8145\\ \hline
BERT-base                                  & 0.7086           & 0.6989\\ \hline
MLP-based Model                            & 0.8406            & 0.8289\\ \hline
\end{tabular}
\end{table}

The BioASQ dataset's results reveal the performance across different model configurations, including BioBERT, AoA Reader, BERT-base, and an MLP-based model, evaluated on accuracy and F1-score (see Table \ref{tab:result2}). The standalone BioBERT model achieves an accuracy of 73.29\% and an F1-score of 72.15\%, establishing a baseline for single-model effectiveness. Combining BioBERT with the AoA Reader significantly enhances performance, with accuracy rising to 82.41\% and F1-score to 81.45\%, illustrating the synergistic benefits of integrating AoA Reader’s dual attention mechanism for improved text understanding. In contrast, the BERT-base model shows lower metrics, with an accuracy of 70.86\% and an F1-score of 69.89\%, highlighting its relative inadequacy compared to BioBERT and AoA Reader combinations. The highest performance is achieved by the MLP-based model, with an accuracy of 84.06\% and an F1-score of 82.89\%, indicating the significant advantage of MLPs in handling non-linear problem spaces and optimizing feature integration. This configuration effectively balances precision and recall, demonstrating superior decision-making capabilities over simpler transformer-based models.

\subsubsection{Method 2: Visualization of BERT and AoA Reader}

\begin{figure}[t]
\centering
\includegraphics[width=0.9\columnwidth]{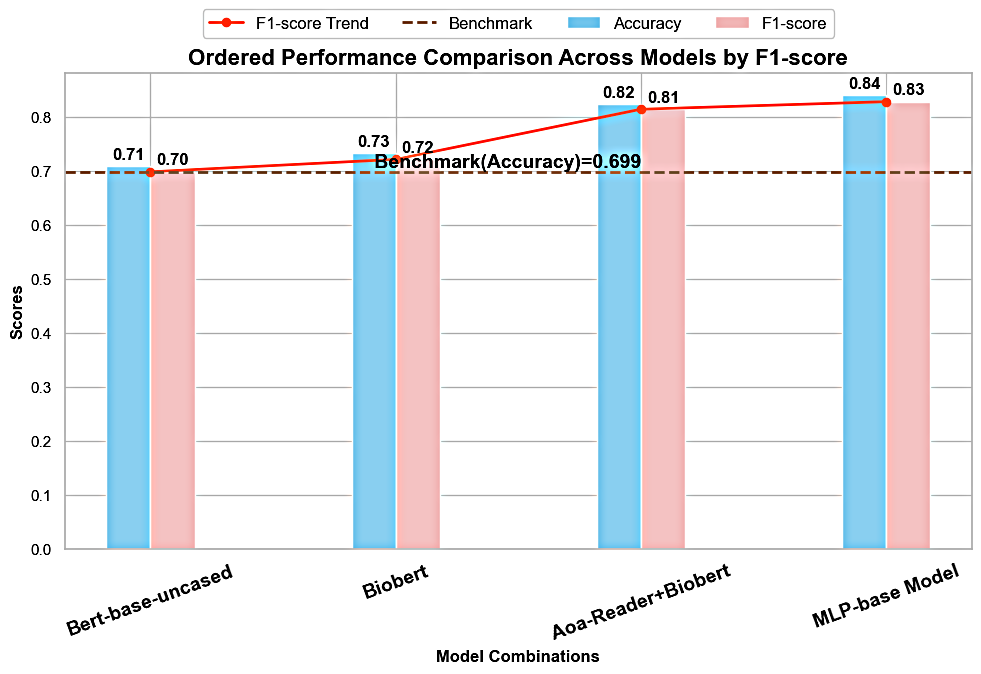} 
\caption{Line chart of BERT and AoA reader}
\label{fig14}  
\end{figure}

The visualization in Figure \ref{fig14} offers a clear comparison of performance metrics across various models, emphasizing both accuracy and F1-scores against a benchmark accuracy of 0.699. The BERT-base model, with an F1-score of 0.71 and an accuracy of 0.70, performs adequately but does not significantly surpass the baseline, suggesting the need for further optimization. The BioBERT model shows a slight improvement in accuracy to 0.73 but a minor drop in F1-score to 0.72 compared to BERT-base, indicating marginal enhancements. Notably, the combination of AoA Reader with BioBERT achieves substantial performance gains, with accuracy and F1-score rising to 0.82 and 0.81, respectively, highlighting the effective synergy of these models in enhancing contextual understanding. The MLP-based model outperforms all configurations, with the highest accuracy of 0.84 and F1-score of 0.83, demonstrating the significant benefits of integrating advanced neural network structures like MLPs. These results underscore the importance of combining multiple technologies for higher accuracy and balanced performance, validating the superiority of integrated and complex model configurations for advanced QA services.

\section{Conclusion}

This study set out to create powerful question-answering (QA) services designed to support healthcare professionals by processing complex biomedical information. We used BERT and BioBERT models, but instead of relying on a single model, we found that integrating multiple models through a multi-layer perceptron (MLP) approach greatly enhanced the system’s performance. By working with datasets like BioASQ and BioMRC, we were able to show that this integration not only improved accuracy but also made the system more adaptable and effective in diverse medical scenarios. A key breakthrough was freezing one BERT model while training another, which prevented overfitting and allowed the system to specialize in certain tasks. The MLP layer played a vital role in synthesizing and prioritizing data, which could be critical for healthcare professionals making high-stakes decisions. These innovations are about more than just technical improvements—they are part of a broader mission to create tools that genuinely support healthcare providers, helping them make faster, better-informed decisions that can lead to better patient outcomes. This work highlights how advanced AI systems can contribute to the common good by making complex medical data more manageable and actionable, ensuring that technology works to benefit everyone in society.

\bibliographystyle{ieeetr}
\bibliography{ref}

\end{document}